\documentclass[sigconf]{acmart}

\usepackage{colortbl}

\usepackage{xcolor}

\usepackage{float} 

\usepackage{booktabs}
\usepackage{amsmath}

\usepackage{hyperref}
\hypersetup{
   colorlinks=true,
   linkcolor=blue,
   filecolor=blue,
   urlcolor=blue,
   citecolor=blue,
}

\usepackage[most]{tcolorbox}

\title{\textbf{AI in Investment Analysis: LLMs for Equity Stock Ratings}}

\author{Kassiani Papasotiriou}
\email{kassiani.papasotiriou@jpmorgan.com}
\affiliation{%
  \institution{J.P. Morgan AI Research}
  \city{New York}
  \country{USA}
}

\author{Srijan Sood}
\email{srijan.sood@jpmorgan.com}
\affiliation{%
  \institution{J.P. Morgan AI Research}
  \city{New York}
  \country{USA}
}

\author{Shayleen Reynolds}
\email{shayleen.reynolds@jpmorgan.com}
\affiliation{%
  \institution{J.P. Morgan AI Research}
  \city{New York}
  \country{USA}
}

\author{Tucker Balch}
\email{tucker.balch@emory.edu}
\affiliation{%
  \institution{Emory University}
  \city{Atlanta}
  \country{USA}
}

\copyrightyear{2024} 
\acmYear{2024} 
\setcopyright{acmlicensed}\acmConference[ICAIF '24]{5th ACM International Conference on AI in Finance}{November 14--17, 2024}{Brooklyn, NY, USA}
\acmBooktitle{5th ACM International Conference on AI in Finance (ICAIF '24), November 14--17, 2024, Brooklyn, NY, USA}
\acmDOI{10.1145/3677052.3698694}
\acmISBN{979-8-4007-1081-0/24/11}

\begin{document}

\begin{abstract}
Investment Analysis is a cornerstone of the Financial Services industry. The rapid integration of advanced machine learning techniques, particularly Large Language Models (LLMs), offers opportunities to enhance the equity stock rating process. This paper explores the application of LLMs to predict stock performance and generate stock ratings by ingesting diverse datasets. Traditional stock rating methods rely heavily on the expertise of financial analysts, and face several challenges such as data overload, inconsistencies in filings, and delayed reactions to market events. Our study addresses these issues by leveraging LLMs to improve the accuracy and consistency of stock ratings. Additionally, we assess the efficacy of using different data modalities with LLMs for the financial domain.\\

We utilize varied datasets comprising fundamental financial, market, and news data from January 2022 to June 2024, along with GPT-4-32k (v0613) (with a training cutoff in Sep. 2021 to prevent information leakage). Our results show that our benchmark method outperforms traditional stock rating methods when assessed by forward returns. Specifically, incorporating financial fundamentals enhances ratings accuracy. While integrating news data improves short-term performance, substituting detailed news summaries with sentiment scores reduces token use without loss of performance. In many cases, omitting news data entirely enhances performance by reducing bias.\\

Our research shows that LLMs can be leveraged to effectively utilize large amounts of multimodal financial data, as showcased by their effectiveness at the stock rating prediction task. Our work provides a reproducible framework for generating consistent and accurate stock ratings, offering a cost-effective and efficient alternative to traditional methods. Future work will extend the analysis to longer time horizons, incorporating more diverse data, and utilizing newer models to enhance detailed investment analysis and reports.
\end{abstract}
\maketitle

\section*{Disclaimer}
This paper was prepared for informational purposes by the Artificial Intelligence Research group of JPMorgan Chase \& Co. and its affiliates (``JP Morgan'') and is not a product of the Research Department of JP Morgan. JP Morgan makes no representation and warranty whatsoever and disclaims all liability, for the completeness, accuracy or reliability of the information contained herein. This document is not intended as investment research or investment advice, or a recommendation, offer or solicitation for the purchase or sale of any security, financial instrument, financial product or service, or to be used in any way for evaluating the merits of participating in any transaction, and shall not constitute a solicitation under any jurisdiction or to any person, if such solicitation under such jurisdiction or to such person would be unlawful.

\section{Introduction}
Investment Analysis is a foundational segment of the financial services industry, and is crucial for the functioning of financial markets, providing essential insights that drive investment decisions, market trends, and economic policies. Financial analysts play a key role in this process by evaluating financial data, preparing reports, and publishing stock ratings among other financial anlysis tasks. Their expertise helps in valuing assets, assessing investment opportunities, and formulating business decisions. By interpreting complex financial information, their analyses help mitigate risks and identify opportunities for investors \cite{fernandez2001role}.

A crucial task of a financial analyst is to publish stock ratings, which evaluate a company's future performance based on forward projections of a company's fundamentals, including earnings, revenue growth, and cash flow, as well as broader market conditions and economic trends. They consist of analysts' expert recommendations on how to position companies over the next quarter to a year and thus play a pivotal role in shaping market perceptions  \cite{singh2021feature}. These ratings are among many variables used to evaluate companies in the investment analysis domain.

In recent years, these methods have been complemented by advanced machine learning techniques, such as Deep Learning methods and Large Language Models (LLMs) \cite{wu2023bloomberggpt, kolte2022stock}. LLMs, with their zero-shot and few-shot learning capabilities, can perform a wide range of tasks without costly fine-tuning. They offer advanced reasoning capabilities and can efficiently handle large volumes of diverse unstructured data, making them useful in financial analysis. Specifically, LLMs can answer questions, summarize information, write content, and handle multiple tasks simultaneously, augmenting the workflow of financial analysts. They significantly enhance the process of generating stock ratings by analyzing financial reports, assessing the sentiment of news articles, evaluating market commentaries, and more.

Predicting stock ratings is a challenging task due to the complexity and dynamic nature of financial markets, but there are several opportunities where LLMs can help: 1) Analysts must process and analyze vast volumes of complex data, and LLMs can assist by efficiently ingesting large datasets. 2) LLMs can generate predictions on demand. 3) LLMs can incorporate information from multiple data sources and modalities which can help reduce bias \cite{singh2021feature,loh2011analyst}.

This study employs an instruction-based general purpose LLM for predicting stock ratings, an under-explored area in research. We utilize various data types such as fundamental financial data (tabular/semi-structured), market data (timeseries), and news data (unstructured), from January 2022 to June 2024. We specifically use GPT-4-32k (v0613), trained on data up until September 2021  \cite{openai2023gpt4_32k} to avoid information leakage. It is important to note that future data leakage is not often accounted for in studies that utilize AI in the financial services domain, which can lead to overly optimistic performance results.

Our method explores best practices for enabling LLMs to utilize various modalities of financial data, and demonstrates its effectiveness at stock rating predictions. Since we do not fine-tune the model, the process remains cost-effective. When evaluated based on forward returns, our method outperforms financial analysts in  our experiments. This highlights the potential of utilizing LLMs in the financial analysis pipeline.

Our key contributions are as follows:
\begin{enumerate}
    \item We demonstrate the novel use of an LLM for predicting stock ratings, addressing a significantly under-researched area.
    \item We evaluate the performance of LLMs in this task to determine which data sources enhance or impede predictive capabilities over various time horizons.
    \item We build a reproducible framework to quickly and consistently generate ratings evaluated by forward-returns.
    \item We address both the prediction of stock ratings and subsequent stock price movements, providing a comprehensive view of utilizing LLMs in the financial forecasting process.
\end{enumerate}
\section{Background and Related Work}

In this section, we first focus on the powerful capabilities of general purpose LLMs in finance and how AI has been increasingly integrated in the domain. We speak specifically to tasks within the area of investing, discussing the importance of AI in financial analysis in conducting these tasks. Finally, we discuss the ratings produced by financial analysts, their impact on the market and the various ways AI has been integrated to enhance the process. 

\subsection{LLMS in Finance} \label{sec:llms_in_finance}
Zero-shot and few-shot LLM techniques are widely applied in the domain of finance. For example, \cite{zhang2023enhancing,dolphin2024extracting} demonstrate using LLMs for identifying sentiment and summarization of financial news, respectively, using instruction based prompting techniques. \cite{srivastava2024assessing} evaluates complex question answering (QA) techniques on semi-structured financial documents. \cite{lee2024survey} assesses LLM performance (ranging from general purpose LLMs to fine-tuned) on QA and summarization for financial documents, text classification, generation, stock movement prediction and more, demonstrating many applications for LLMs in finance. 

The commonalities among research utilizing LLMs for prediction include using diverse datasets, employing LLMs at multiple stages and enhancing interpretability. The integration of the capabilities highlighted above and advancements in LLMs significantly enhance financial tasks such as stock movement prediction, risk mitigation and quantitative trading. \cite{lopezlira2023chatgptforecaststockprice} uses GPT to predict stock market returns from news headline sentiment scores and finds a positive correlation, outperforming older GPTs and BERT in forecasting returns, as evaluated by the Sharpe Ratio. \cite{fatouros2024largelanguagemodelsbeat},  \cite{tong2024ploutosinterpretablestockmovement}, and \cite{li2024alphafinbenchmarkingfinancialanalysis} utilize several sources of data such as financial news, fundamentals, stock prices, market data and macroeconomic factors to aid in stock prediction. \cite{fatouros2024largelanguagemodelsbeat} applies Chain-of-Thought (CoT) prompting and In-Context Learning with GPT-4 to generate signals and subsequently ranking strategies that show positive percentage returns on selected stocks. \cite{cao2024risklabspredictingfinancialrisk} leverages LLMs to analyze and predict financial risks by combining data from earnings calls, market-related time series data, and contextual news data.  For quantitative trading, a popular avenue of research utilizes memory modules and knowledge bases to aid in a model's self-adaptability. \cite{yu2024finmem,li2023tradinggpt} built LLM-based autonomous trading agents that utilize layered memory. \cite{wang2024quantagent} built a self-improving LLM using an agent that refines its responses with a knowledge base and then tests responses in real-world scenarios to update the knowledge base with new insights. 
\subsection{Analyst Stock Ratings} \label{sec:analyst_stock_ratings}
An analyst stock rating forecasts a stocks performance. In the most common scenario, analysts publish ratings upon the release of quarterly filings, earnings calls, or significant events, updating their guidance for the next quarter, and for rest of the year. These ratings fall into five categories, though terminology may vary: 

\begin{itemize}
    \item \textbf{Strong Buy}/\textbf{Buy}: Indicates that the stock is expected to significantly outperform the market or its sector.
    \item \textbf{Moderate Buy}:  Suggests that the stock is expected to perform better than the market average or its sector. Also referred to as \textbf{Outperform} or \textbf{Overweight}.
    \item \textbf{Hold}: Indicates that the stock is expected to perform in line with the market or its sector.
    \item \textbf{Moderate Sell}: Suggests that the stock is expected to perform worse than the market average or its sector. Also referred to as \textbf{Underperform} or \textbf{Underweight}.
    \item \textbf{Strong Sell}/\textbf{Sell}: Indicates that the stock is expected to underperform its benchmark significantly.
\end{itemize}

Different institutions utilize custom rating scales, which can vary between organizations. For example, some analysts use a one-to-five rating system based on risk-adjusted performance, others employ a buy, hold, sell system, while certain firms use proprietary aggregated scores that combine ratings from multiple research providers. These rating scales are useful because they provide investors with tailored insights and help them make informed decisions based on different analytical perspectives.

Analysts utilize past and current qualitative and quantitative information about a company's performance, \cite{zaremba2015profitability, womack1996brokerage} and based on these evaluations and their experience, they recommend stock ratings that are used by investors to make decisions regarding an asset \cite{fernandez2001role,loh2011analyst}.
Common data includes: 1) Fundamental and technical analysis: Assesses intrinsic value and trading trends using metrics like earnings-per-share, shares outstanding, return on assets, price, volume, and more.  2) Company and sector news: Considers management changes, product launches, mergers, and regulatory developments. 3) Market and sector performance: Evaluates overall market trends, sector performance, and price movements \cite{zaremba2015profitability,jimenez2021counselors,fernandez2001role}.

\subsection{Importance of Stock Ratings} \label{sec:ai_for_analyst_stock_ratings}
As mentioned in section \ref{sec:llms_in_finance}, while research on company performance prediction shows promising results, it often overlooks stock ratings, a key indicator of future stock performance. Investors use stock ratings for many tasks such as portfolio building, risk assessment, asset allocation and other investment strategies\cite{jegadeesh2004analyzing, sood2023deep, brown2014analyst}. Studies on how ratings impact the market reveal that investors closely monitor these ratings to make informed decisions, leading to market movements based on ratings. For example, \cite{singh2021feature} analyzes 20 years of S\&P500 trading data, developing a classifier that predicts 1\% price changes up to 10 days ahead with 83.62\% accuracy, 85\% precision for buy signals, and 100\% recall for sell signals. Feature ranking highlights analyst stock ratings as top contributors.  \cite{jegadeesh2006value} evaluates the impact of analysts' recommendations in G7 countries, finding significant stock price reactions to recommendation revisions in all countries except Italy, with the US showing the largest price reactions and post-revision price drifts. \cite{barber2010ratings} evaluates the impact of analyst stock changes, finding that influential recommendations are linked to increased stock volatility and significant changes in consensus earnings forecasts. \cite{zaremba2015profitability} examines the profitability of analysts' recommendations in the Polish market using data from 2004-2013. It builds market-neutral portfolios and tests their performance against CAPM, Fama-French, and Carhart models. The key finding is that strategies based on analysts' recommendations yield statistically significant positive abnormal returns, demonstrating their profitability. 

\section{Methodology} \label{sec:methodology}

We leverage LLMs to analyze financial data and generate stock ratings, taking advantage of their ability to process large volumes of information, recognize complex patterns, and adapt to new data. LLMs can efficiently handle diverse data sources and provide detailed insights that traditional methods might miss. Our goal is to provide the LLM with the same information an analyst would consider, such as financial fundamentals, stock price movements, news summaries, sentiment, and other relevant data. This helps us assess the feasibility of using LLMs for investment analysis and identify the techniques and information that improve their performance.

\subsection{Prompt Structure} \label{sec:prompt_structure}
We utilize GPT-4-32k (v0613), hosted on Azure, which features a context window of 32,000 tokens and is trained on data up to September 2021 \cite{openai2023gpt4_32k}. We specifically selected this model to prevent information leakage, as the data we use is from after the model's training cut-off date. 

We use the system prompt to instruct the LLM to adopt the persona of a financial analyst. By defining this role, we provide the LLM with a clear framework for its function. Additionally, we contextualize the financial terms utilized in the experiments by defining the scale of  stock ratings and their definitions (Section~\ref{sec:analyst_stock_ratings}), incorporating synonyms to account for variations in terminology. We also provide detailed descriptions of financial fundamentals, which are outlined in \ref{sec:data}.

To design the user prompt, we follow the success of Chain-of-Thought and few-shot prompt approaches \cite{wei2022chain, brown2020language}, and encourage the LLM to engage in reasoning before making its final prediction and provide it with an example of what the output should look like. Additionally, we provide company-specific input data in a structured format, with textual information first followed by numerical data in tables, following findings from \cite{sui2024table}. We also perform basic CoVE (chain of verification) to detect if it's is predicting things for the correct dates.

\subsection{Problem Formulation} \label{sec:problem_formulation}

Let \( \text{rating}_{c}(t, p) \) be a rating for a company \( c \) released on date \( t \), predicting the company's performance at a future horizon of \( p \) months. The rating can take any of the following ordinal values:
\[
\text{rating}_{c}(t, p) \in \{-2, -1, 0, 1, 2\}
\]
where -2 = Strong Sell, -1 = Moderate Sell, 0 = Hold, 1 = Moderate Buy, and 2 = Strong Buy.

We assess the accuracy of a rating by evaluating how the company's stock performs. This approach is commonly used in technical papers that utilize future returns. For example, to assess the value of analyst stock ratings  \cite{jafari2022gcnet} investigates the performance of stocks shortly after ratings are published, while \cite{barber2006buys} analyzes the distribution of analyst stock ratings over time, using company returns in quintiles to predict potential profitability. \cite{barber2010ratings} evaluates changes in analyst stock ratings by comparing company returns across different rating levels.

To determine the accuracy of a rating \( \text{rating}_{c}(t, p) \), we evaluate the performance of company \( c \) using its forward returns (at period \( t + p \)),  and compare it to other companies. This is done by computing company returns for the entire group (e.g. S\&P500 constituents) at a fixed time horizon, and then dividing these into quintiles. The quintile groups correspond to rating levels, e.g., companies with returns in the lowest quintile significantly underperformed their peers, making their ground truth rating a Strong Sell.

Our process is as follows. We first calculate forward company returns as well as market and sector relative returns:

Given the price for company $c$ at time $t$, $P_c(t)$, the company return \( R_c(t, p) \) over the period \( p \) is defined as:
\[
R_c(t, p) = \frac{P_c(t + p) - P_c(t)}{P_c(t)}
\]

The sector-relative forward return \( R_{c,s}(t, p) \) is defined as:
\[
R_{c,s}(t, p) = R_c(t, p) - R_s(t, p)
\]
where the sector return \( R_s(t, p) \) over the same period \( p \) is:
\[
R_s(t, p) = \frac{P_s(t + p) - P_s(t)}{P_s(t)}
\]
For a rating released on date \( t \) with a horizon of \( p \), we compute the quantiles of returns across all companies \( c \) at \( t + p \) and assign each company the quantile that their returns fall into. If the returns quantile matches the rating, then the rating is considered correct.

Let \( Q_c(t, p) \) represent the quantile of the company returns \( R_c(t, p) \), \( Q_{c,m}(t, p) \) represent the quantile of the market-relative returns \( R_{c,m}(t, p) \), and \( Q_{c,s}(t, p) \) represent the quantile of the sector-relative returns \( R_{c,s}(t, p) \).

We define the indicator function for the correctness of the rating \( \text{rating}_{c}(t, p) \) with respect to the absolute performance quantile \( Q_c(t, p) \) as follows:
\[
\mathbb{I}(\text{rating}_{c}(t, p) = Q_c(t, p)) = 
\begin{cases} 
1 & \text{if } \text{rating}_{c}(t, p) = Q_c(t, p) \\
0 & \text{otherwise}
\end{cases}
\]
Similarly, for sector-relative returns:
\[
\mathbb{I}_{\text{sector}}(\text{rating}_{c}(t, p) = Q_{c,s}(t, p)) = 
\begin{cases} 
1 & \text{if } \text{rating}_{c}(t, p) = Q_{c,s}(t, p) \\
0 & \text{otherwise}
\end{cases}
\]
\section{Experiments} \label{sec:experiments}

In this section we provide an overview of the data we use as well as our experiment set up.

\subsection{Data} \label{sec:data}
Our analysis focuses on US equities, specifically the 500 constituents of the S\&P 500 index, using data spanning from January 2022 to the end of June 2024.  

\textbf{Analyst Stock Ratings}:
We gather analyst stock ratings for each company in the S\&P500 \cite{yahoo_finance_api}. Out of a total of 45,000 ratings from 126 firms, the majority of ratings (75.90\%) were maintained, followed by reiterate (7.25\%), downgrade (6.27\%), upgrade (5.68\%), and initiate (4.89\%). The top five firms, which include Morgan Stanley, Barclays, Wells Fargo, Citigroup, and RBC Capital, collectively account for 31.61\% of all ratings. Specifically, Morgan Stanley contributed 9.99\%, Barclays 6.52\%, Wells Fargo 5.91\%, Citigroup 4.67\%, and RBC Capital 4.52\% of the total ratings. This dataset comprises the firm issuing the rating, the date of the rating, and the rating itself.  However, we do not have data for the target date or the target price. Note that for a particular date and company, there may be multiple ratings issued by different firms. 

\textbf{Financial News Summaries}:
We collect news articles for stocks in the S\&P500 \cite{yahoo_finance_api} \cite{serpapi} \cite{langchain}. We filter irrelevant content out by performing named entity recognition (NER), utilizing both company names and possible company aliases to enhance this process (alias were also scraped from \cite{yahoo_finance_api}). After filtering down to relevant data, our dataset consists of the following: on average per month, there are 39.63 articles, 187K characters, and 40K tokens, with 74.70 URLs and 34.40 missing articles per ticker. We summarize the monthly news for each company and sector using GPT-4-32k (v0613) to highlight key events and trends. Our system prompt designates the user as an expert news summarizer, tasked with condensing articles into concise summaries that highlight key events and important information, while excluding irrelevant content. We utilize two user prompts, one to create summaries for both individual companies, and another to summarize news across an entire sector, identifying general themes and trends. 

\textbf{Financial News Summaries Sentiment}:
We utilize GPT-4-32k (v0613) to identify the sentiment of the financial news summaries on the company and sector level. The system prompt assigns the user the role of an expert in news sentiment scoring, particularly for financial markets, using a scale from -5 to 5 to indicate the sentiment's severity. We utilize two user prompts, one to score the sentiment of news summaries for individual companies, and another for scoring sentiment at the sector level. Each template provides specific instructions and examples to ensure consistency and accuracy in summarization and sentiment scoring. 

\textbf{Historical Returns}:
For a similar time frame as the news summary data, we collect daily stock prices for companies in the S\&P500 and compute technical indicators using the prices \cite{yahoo_finance_api}. The metrics include current price, the 52-week price range, 90-day volatility, and performance metrics over 1-month, 3-month, and 12-month periods. The performance metrics are divided into returns, market relative returns, and sector relative returns. 

\begin{figure}[h!]
    \raggedleft
    \includegraphics[width=1.\linewidth]{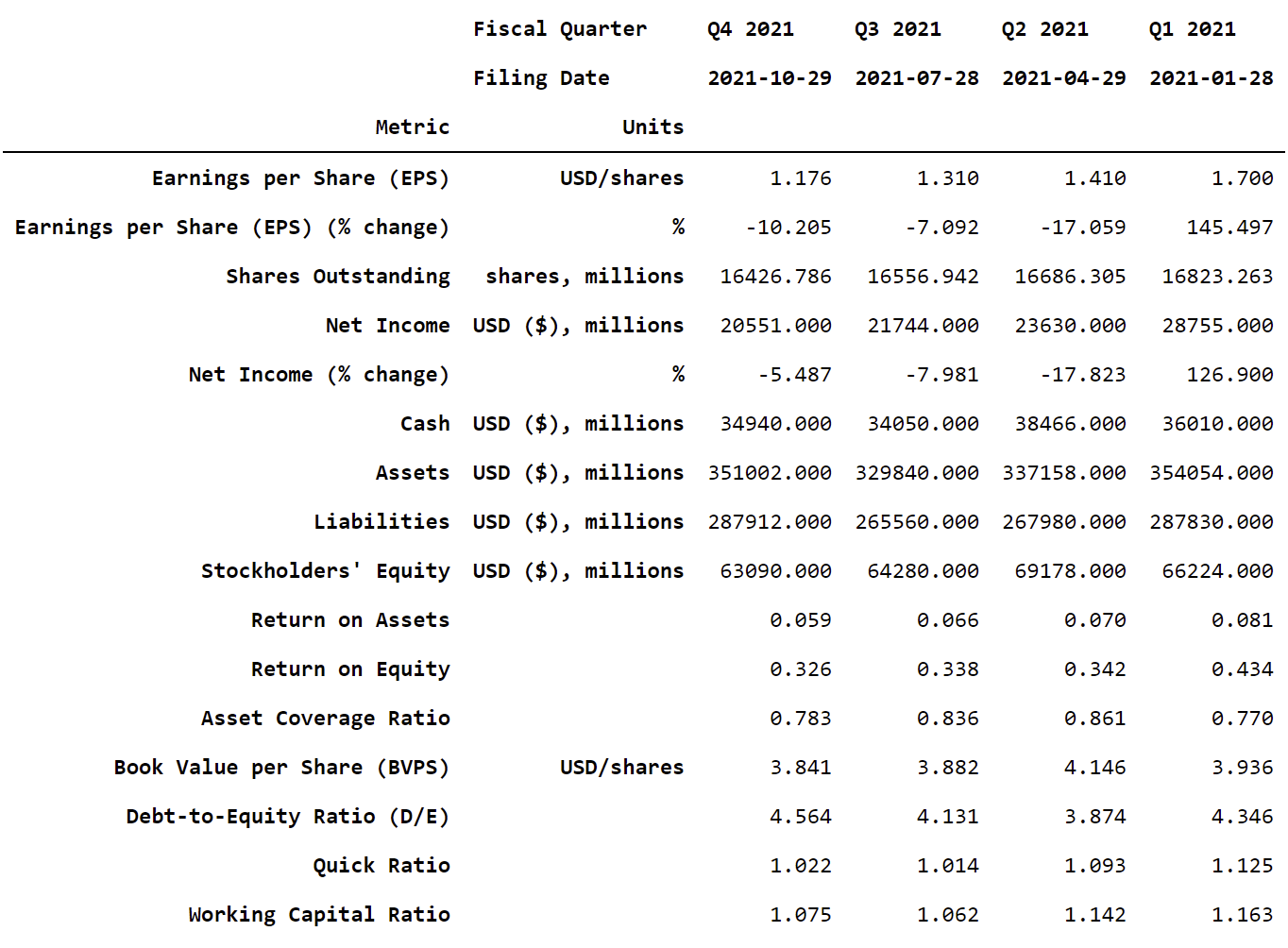}
    \caption{Financial Fundamentals Example: AAPL}
    \label{fig:fin_fundamentals_example_aapl}
\end{figure}
\textbf{Financial Fundamentals}:

We aggregate quarterly company fundamentals from 10-Q and 10-K filings from January 2022 to March 2024 using finagg API \cite{finagg} to access the SEC API \cite{secapi}.  We use the past 4 quarterly fillings available for each prediction date. These filings, submitted by companies to the U.S. Securities and Exchange Commission (SEC), provide detailed information on their balance sheet, income statements, and cash flow statements. Metrics provided to LLM are listed in Figure \ref{fig:fin_fundamentals_example_aapl}.

\subsection{Experiment Setup} \label{sec:experiment_setup}

We utilize GPT-4-32k (v0613), hosted on Azure, which features a context window of 32,000 tokens and is trained on data up to September 2021 \cite{openai2023gpt4_32k}. We specifically selected this model to prevent information leakage, as the data we use is from after the model's training cut-off date.

Each experiment involves asking the LLM to generate stock ratings at the beginning of each month from January 2022 to June 2024 for every company in the S\&P500. Specifically, for each company, we let the LLM know that it needs to issue a rating at the start of a given month. We provide it with additional data (which varies between experiments as described below) and ask it to predict whether the stock should be rated as Strong Sell, Moderate Sell, Hold, Moderate Buy, or Strong Buy, for multiple future time horizons: 1, 3, 6, 12, and 18 months. For example, we ask the LLM to predict the rating it would release in March 2022 for future dates such as April 2022, June 2022, September 2022, March 2023 and September 2023.

For each method, this results in approximately 5 (horizons) * 30 (start dates) * 500 (companies) ratings. Using multiple time horizons allows us to account for the uncertainty around the exact target date of the analyst’s rating, enabling us to assess predictions at several points in the future. Additionally, varying time horizons helps us evaluate the LLM's predictive performance across different periods.

\subsubsection{Methods} \label{sec:methods}
We conduct experiments with five distinct methods: Vanilla, News, Sentiment, Fundamentals, and Fundamentals + Sentiment. All of these methods utilize GPT-4-32k (v0613), but with varying data provided to the model in its input context. We provide the LLM with the task description (as highlighted in Section~\ref{sec:prompt_structure}. For each query, we include the name of the company, the date on which the ratings will be released, and the five forward-looking time horizons for which it needs to generate ratings. Inspired by the chain-of-thought prompting framework, we also ask the LLM to output the corresponding price targets, along with a short explanation. We check the LLM's response to verify that it is computing the dates which each time horizon corresponds to correctly. 

\textbf{Vanilla}: The input context includes a snapshot of the company's historical data: returns, market-relative returns, and sector-relative returns for the past 1-month, 3-month, and 12-month periods. Additionally, we provide the current stock price (as of rating date), the 52-week price range (min, max), and the 90-day volatility (std. dev. of daily returns). In total, the LLM receives 10 values describing historical returns (1 for volatility), plus 3 values relating to the stock price (current and 52-week min-max), for a total of just 13 numbers. We found that these simple data points greatly improve the LLM's ability to generate accurate ratings. This setting serves as our baseline for the following experiments.  

\textbf{News}: This experiment enhances the input prompt for the \textit{Vanilla} method by including news data. As it is not pragmatic to include entire news articles due to LLM context limits, we provide summaries of company news and sector news from the previous month. In addition to the outputs described above, the LLM is also tasked with assessing the sentiment of the news summaries provided (positive, negative, neutral, or mixed), and to use this in its predictions. We found improved performance when the LLM receives the news summaries earlier in the context (before the technical indicators).
Given the success of in-context learning, we also provide the LLM with an example: the input data as described, along with the expected dates, an explanation of the ratings, and the corresponding rating output to guide its predictions. We use the dates as a hallucination check, incorporating a Chain of Verification (CoVE) to ensure that the predicted dates are correct, as they are a straightforward aspect to verify. If the LLM fails to get the dates right, it suggests that correctly predicting the ratings would likely be even more challenging.

\textbf{Sentiment}: This experiment is similar to \textit{Vanilla}, save for one key difference — the inclusion of pre-computed news sentiment (also computed using GPT-4-32k (v0613)). Unlike the \textit{News} experiment, which provides the LLM with descriptive news summaries, this method supplies the LLM with two sentiment scores, one for company news sentiment, and one for sector news sentiment, from the previous month. The summaries used in the sentiment scoring process are the same as the ones provided to the LLM in the \emph{News} method. The sentiment scores range from -5 to 5, capturing a a spectrum of sentiment from extremely negative to extremely positive. 

\textbf{Fundamentals}: This method augments the \textit{Vanilla} prompt with quarterly financial fundamental data. The model is supplied with company financial metrics and detailed descriptions of each metric. The system instructions for the LLM are updated with definitions of the fundamental features. The LLM is tasked with analyzing these numbers in its process. The fundamentals, as show in Figure ~\ref{fig:fin_fundamentals_example_aapl}, are provided to the LLM in an HTML format, as HTML seems outperform other formats for LLM ingestion, as mentioned in Section~\ref{sec:prompt_structure}.

\textbf{Fundamentals + Sentiment}: 
This experiment builds upon the \textit{Fundamentals} method by also including the sentiment scores used in the \textit{Sentiment} method; the setup is similar, with two scores capturing company sentiment and sector sentiment from news from the previous month. The model is prompted to use both the fundamental data and sentiment scores to make its recommendations.

\subsection{Evaluation} \label{sec:evaluation}
We evaluate ratings based on forward returns over 1, 3, 6, 12, and 18-month periods, including evaluations for market-relative and sector-relative returns. As described in Section \ref{sec:problem_formulation}, a rating is considered correct if the quantile for the true forward return aligns with the rating's rank. For example, let's take a rating for a given company (whether from an analyst or LLM), with a 6-month horizon. Suppose the stock was rated as a Strong Buy for the 6-month horizon, and the company's 6-month forward return falls in the bottom quantile (based on 6-month forward returns from the same date for all companies). This constitutes a significantly incorrect rating, as the company was amongst the worst performers in the market, but was rated as a Strong Buy. The ground-truth rating in this case would have been a Strong Sell. Conversely, if another method generated a rating of Hold for the same <company,date,horizon> combination, the rating would still be incorrect, but less severely so.

We compute the \textbf{Mean Absolute Error (MAE)} using two types of returns — regular \textit{market-relative forward returns} (these automatically become market-relative due to our quantile ranking evaluation), and \textit{sector-relative forward returns} (where the subsector's forward return is subtracted from the asset return). MAE is appropriate for ordinal classification because it considers the magnitude of the error and accounts for how far a rating is from its true value. Ratings further away from the ground-truth are penalized more. Accuracy, a popular metric for classification tasks, treats all errors equally, regardless of their severity. Since we have a balanced distribution of classes (due to quantization), MAE doesn't need to be adapted, however metrics such as macro-averaged MAE~\cite{baccianella2009evaluation} can be utilized in those scenarios. 

We compute a \textbf{composite error} to compare methods more easily; we average the marker-relative return based MAE over the three most common time horizons in this domain – 3, 6, and 12-month periods – as the 1-month rating is usually too soon to be useful. We exclude the 18-month predictions from this score because ratings are typically intended for up to one year, and analysts usually update their ratings for longer-term horizons. Table~\ref{tab:evaluation_comparison_all_metrics} presents these values. We also present a monthly breakdown of performance of all methods across 1, 3, 6, 12, and 18-month periods. Figure~\ref{fig:MAE_comparison_bar} displays these results. 

Please note \textbf{Analyst} represents the real-world ratings from financial analysts across various Wall Street firms, which we measure against the LLM's predictions.

\section{Results} \label{sec:results}

\begin{figure*}[htp!]
    \centering
    \includegraphics[width=1\linewidth]{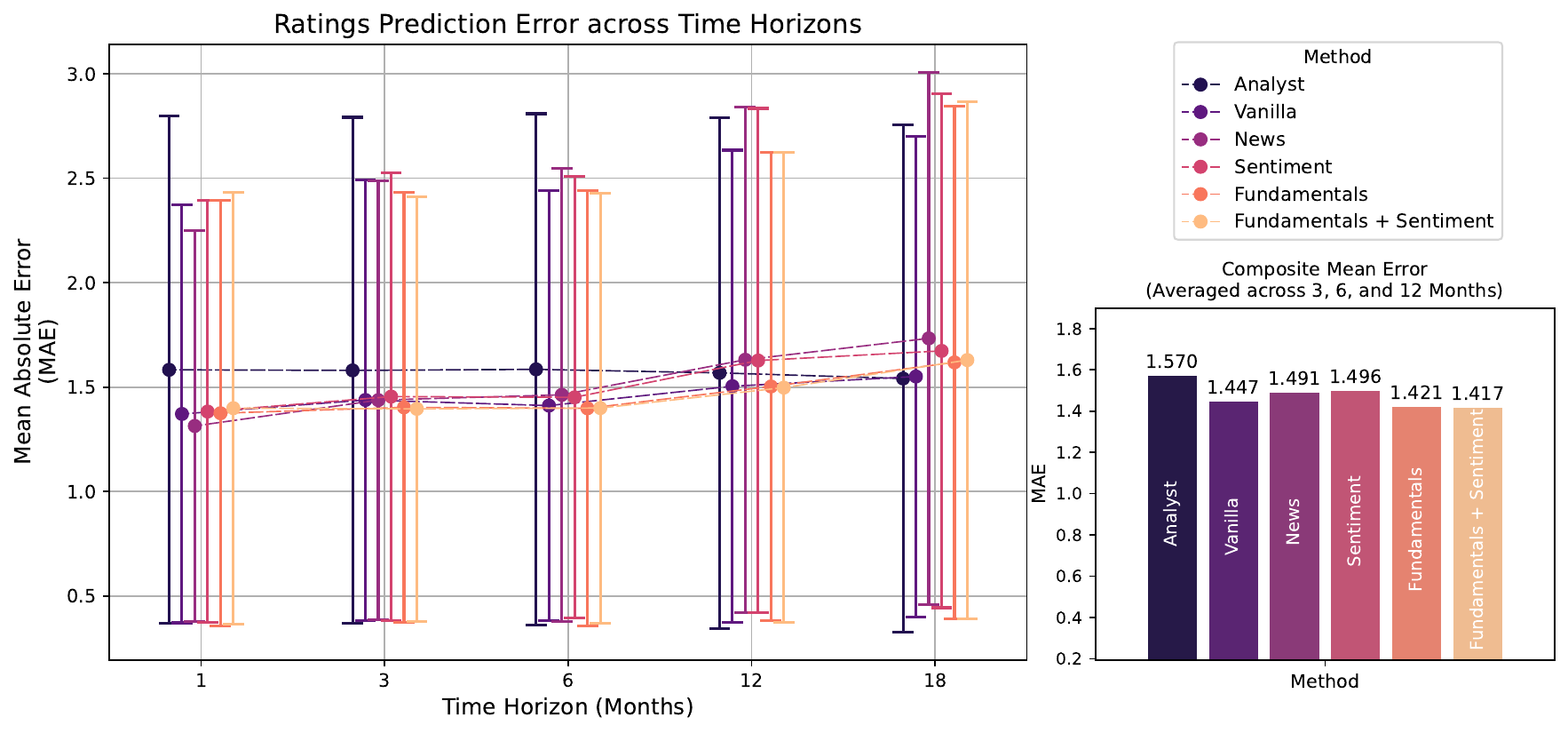}
    \caption{Ratings Prediction Error (MAE) across different time horizons.
 a) Mean, Standard Deviation for each technique and time horizon.  b) Composite Mean Error (MAE averaged across 3,6,12-month time horizons).}
    \label{fig:MAE_comparison_bar}
    \Description[Rating Prediction Error across Time Horizons]{}
\end{figure*}

Our findings from the month-wise breakdown of the \textit{market-relative MAE} and \textit{sector-relative MAE}  across all methods. Figure~\ref{fig:MAE_comparison_bar} visualizes how the methods stack up to each other, including a snapshot comparison using the \textit{composite error} (also listed in Table~\ref{tab:evaluation_comparison_all_metrics}). Note that the figures are based on the \emph{market-relative MAE}. 
\begin{figure}[h!]
    \centering
    \includegraphics[width=1\linewidth]{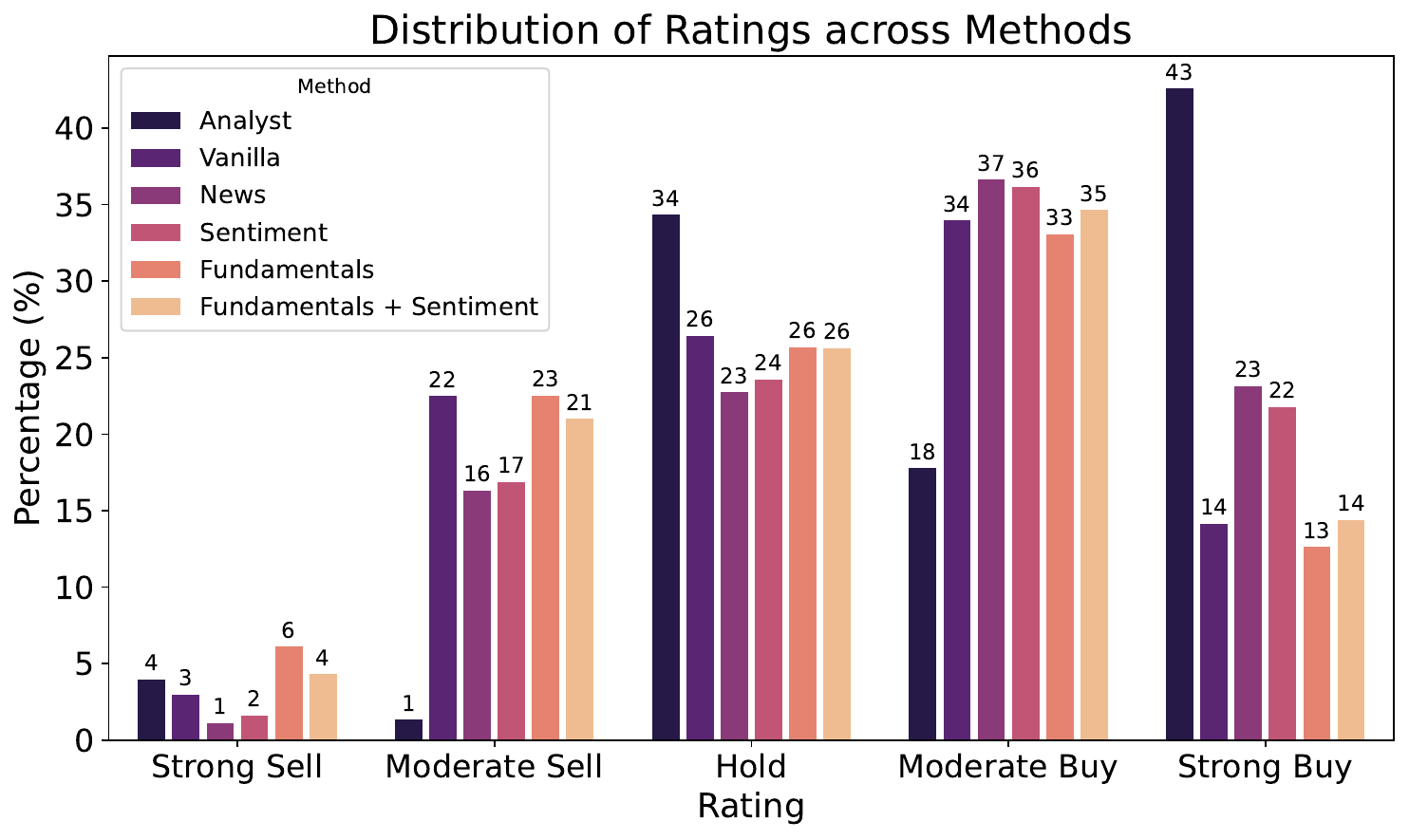}
    \caption{Value Counts of Ratings Across All Months for All Experiments}
    \label{fig:ratings_distributions}
\end{figure}

\begin{table}[h!]
\centering
\scriptsize 
\setlength{\tabcolsep}{4pt}
\begin{tabular}{>{\raggedright\arraybackslash}p{0.8in}>{\centering\arraybackslash}p{0.8in}>{\centering\arraybackslash}p{0.8in}}
\toprule
\textbf{Model} & \textbf{Return MAE ± Std} & \textbf{Sector Rel Return MAE ± Std} \\
\midrule
\textbf{Analyst} & 1.570 ± 0.637 & 1.591 ± 0.648 \\
\textbf{Vanilla} & 1.447 ± 0.745 & 1.459 ± 0.749 \\
\textbf{News}& 1.491 ± 0.738 & 1.513 ± 0.744 \\
\textbf{Sentiment}& 1.496 ± 0.752 & 1.512 ± 0.755 \\
\textbf{Fundamentals} & 1.421 ± 0.732& 1.439 ± 0.739\\
\textbf{Fundamentals + Sentiment}& 1.417 ± 0.747& 1.441 ± 0.752 \\
\bottomrule
\end{tabular}
\caption{Evaluation Averaged Across 3, 6, and 12 Month Periods for Experiments}
\label{tab:evaluation_comparison_all_metrics}
\end{table}

\subsection{Traditional Analyst vs. Vanilla LLM} \label{sec:analyst_vs_vanilla} 
Figure~\ref{fig:ratings_distributions} shows how analysts are heavily biased towards buy ratings, and only gave sell ratings less than 5\% of the time. In Table~\ref{tab:evaluation_comparison_all_metrics}, the \textit{Vanilla} method has a lower MAE of 1.447 compared to the Analyst predictions, which has a Return MAE of 1.570. This indicates that the LLMs predictions, even with only basic financial data, are more accurate than those made by analysts. However, the standard deviation of the \textit{Vanilla} method is 0.745, higher than the Analyst's 0.637, suggesting that while the predictions are more consistent, they are less accurate overall. Sector Return MAE and standard deviation follow the same trend. Figure~\ref{fig:MAE_comparison_bar} shows that errors for the Analyst predictions decrease as the look-ahead periods increase, with slightly better performance in the 18-month period, while errors for \textit{Vanilla} experiment increase.

\subsection{News: Summary vs. Sentiment} \label{sec:news_vs_sentiment}
Table~\ref{tab:evaluation_comparison_all_metrics} shows that the \textit{News (Summary)} experiment, which we provide the previous month's news summaries for the company and the sector, results with a Return MAE of 1.491 and a standard deviation of 0.738. The \textit{News (Sentiment)} experiment, which we provide sentiment scores of the news summaries instead of summaries (scored on a scale of -5 to 5), results in a Return MAE of 1.496 and a standard deviation of 0.752. Interestingly, neither outperformed the \textit{Vanilla} experiment. Additionally, we did not see improved performance when including summaries compared to only including their sentiment. The trends for Sector Relative Return MAE are consistent with the Return MAE metrics. Figure~\ref{fig:MAE_comparison_bar} shows that the \textit{News (Summary)} experiment performs best in the 1-month period, outperforming all other experiments in both Return and Sector MAE. This suggests news summaries provide better short-term predictions, likely because we include summaries from the previous month, therefore offering a clearer picture of recent company performance. The \textit{News (Sentiment)} experiment performs similarly to the \textit{News (Summary)} experiment, indicating that incorporating sentiment does not significantly improve performance compared to news summaries.

\subsection{Fundamentals vs. Fundamentals + Sentiment} \label{sec:fundamentals_vs_fundamentals_sentiment}
Table~\ref{tab:evaluation_comparison_all_metrics} shows that the \textit{Fundamentals + Sentiment} experiment has the best performance in terms of Return MAE, with a value of 1.417, indicating the most accurate predictions. The \textit{Fundamentals} experiment has a Return MAE of 1.421 and a lower standard deviation of 0.732, indicating more consistent predictions.

Figure~\ref{fig:MAE_comparison_bar} shows that both the \textit{Fundamentals} and \textit{Fundamentals + Sentiment} experiments consistently perform best across most months, particularly excelling in the 3, 6, and 12-month periods. This stable performance across horizons reinforces the benefits of incorporating fundamental financial data. The \textit{Fundamentals + Sentiment} experiment outperforms in the 3 and 6-month periods, demonstrating that combining fundamentals and sentiment scores is effective in the short term but may lead to conflicting signals over longer periods, as indicated by the higher MAE in the 18-month period. Both models outperform the \textit{Vanilla} experiment and Analyst predictions, highlighting the significant impact of financial fundamentals. Including company and sector sentiment, without the actual news summary data, improves prediction accuracy, likely due to decreased complexity and noise from news data.

\subsection{Results Summary}
Overall, for all LLM experiments \textit{(Vanilla, News (Summary), News (Sentiment), Fundamentals and Fundamentals + Sentiment)}, the errors increase as we make predictions further into the future, indicating that the LLMs are better at short-term predictions and struggle with longer-term forecasts. News-based experiments (especially \textit{News (Summary))} perform best in the short term due to the immediate impact of news. We find that the \textit{News (Sentiment)} experiment generally performs similarly to the \textit{News (Summary)} experiment, indicating that incorporating sentiment analysis does not significantly improve performance compared to providing news summaries. \textit{Fundamentals} and \textit{Fundamentals + Sentiment} experiments also perform similarly,  excelling in the medium term. Finally, Analyst predictions show the best performance over longer periods. 

\subsection{Efficacy of News}
\begin{figure}[h]
    \centering
    \includegraphics[width=1\linewidth]{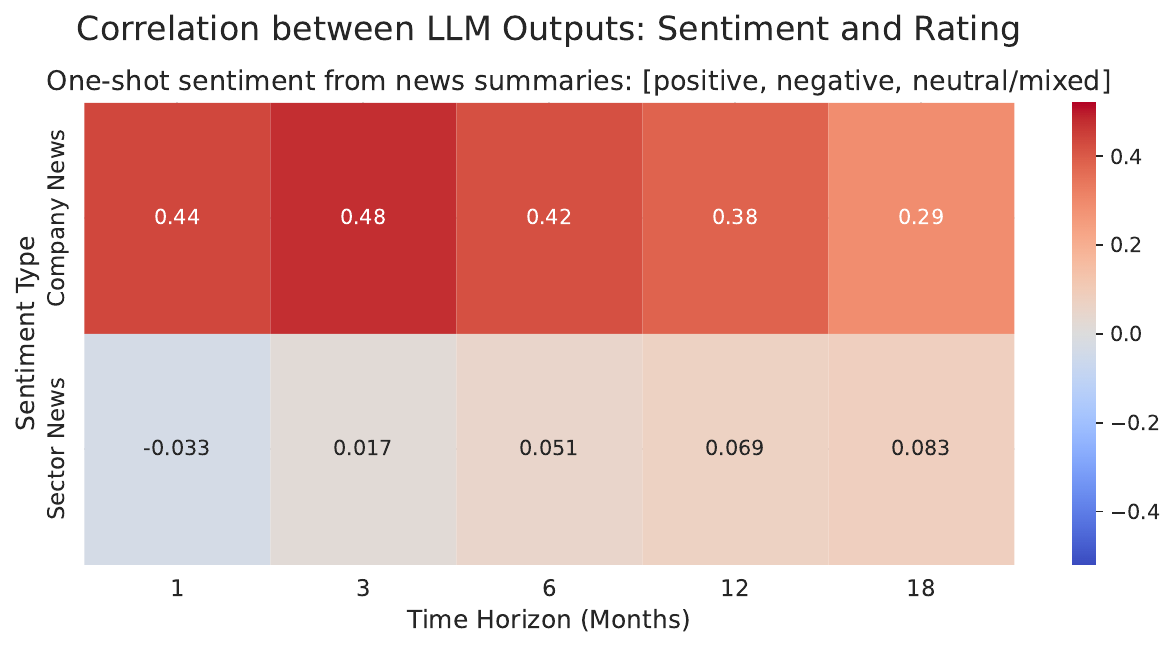}
    \includegraphics[width=1\linewidth]{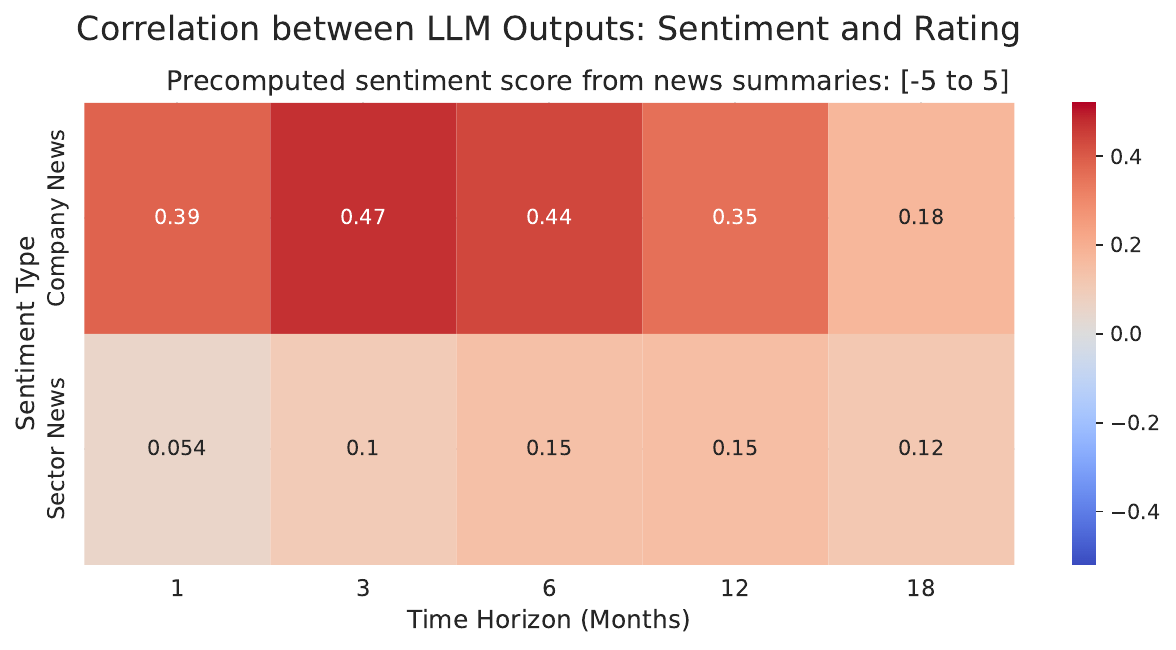}
    \caption{Correlation between LLM's sentiment prediction and ratings across time horizons for two methods: \emph{News} (top) and \emph{Sentiment} (bottom).}
    \label{fig:news_ratings_correlation}
\end{figure}

To understand the impact of news summaries on the results, we compute the Spearman correlation and generate heatmaps for the news summaries and news sentiment. In the \textit{News} experiment, we ask the LLM to provide a rating for the company news summary and the sector summary before predicting the stock ratings. For the \textit{Sentiment} experiment, we score each sector and news summary for its sentiment and then provide these sentiment scores during inference instead of the news summaries. In both cases, we observe that news summaries are correlated across months, especially with the periods closer to the rating. Additionally, the heatmaps Figure ~\ref{fig:news_ratings_correlation} reveal that LLM ratings are correlated with the predictions made for the previous period. Moreover, Figure~\ref{fig:ratings_distributions} show how utilizing news-derived data biases the model to make more positive ratings.

Additionally, Figure ~\ref{fig:news_ratings_correlation} shows a strong positive correlation between the LLM's ratings and the sentiment score derived from the news summaries, indicating that more positive sentiment leads to more favorable LLM ratings. This influence of news sentiment is reflected in the distribution of ratings, where we observe an increase in positive ratings, contributing to less accurate ratings.

\subsection{Challenges and Limitations}
One limitation of this study is our method of evaluating ratings, which is based on forward returns over fixed periods and the quantiles into which these returns fall. These returns could be sensitive to market conditions, which might experience abnormal shifts on specific days, thereby affecting the evaluation. Additionally, there are other factors that determine if the ratings were correct or not, which might be more qualitative than quantitative, such as market sentiment, company-specific news, and broader economic indicators. Moreover, our approach to evaluating the analysts was tricky since we did not have the exact target date for the ratings (hence an assessment with varied time horizons). Another challenge is that we did not provide the model with many essential factors that analysts consider, such as projections of future performance, earnings call reports, investor sentiment, and other qualitative assessments. Additionally, we did not test the model's ability to process and understand extremely large amounts of information, which analysts often review in their evaluations.

\section{Conclusion}
This study explores the potential of LLMs to predict stock ratings, a novel application within the finance sector. By integrating various types of information, including basic financial metrics, technical indicators, financial news summaries financial news sentiment, and financial fundamentals, we aim to evaluate the performance of LLMs in this task and understand which data sources enhance or hinder their predictive capabilities.

\textbf{Key Findings:}\begin{enumerate}
    \item     The benchmark Vanilla LLM model, which uses basic financial metrics, demonstrates stronger performance than traditional analyst evaluations when assessed by forward returns.
    \item The Fundamental LLMs outperformed all experiments, highlighting the significant impact of financial fundamentals on prediction accuracy. Additionally, combining sentiment scores with this data, without the full news summaries, further improved prediction accuracy.
    \item Integrating news summaries and sentiment analysis provides some short-term predictive benefits but does not significantly improve long-term prediction accuracy when compared to the Vanilla model.
    \item The performance difference between adding news as text versus news sentiment to the LLM is very small when other data is not included (i.e. Fundamentals), indicating that both approaches offer similar benefits.
    \item LLMs perform better in short-term predictions, which encourages further exploration of their capabilities for shorter period company predictions.
    \item News summaries are more beneficial for short-term predictions, while traditional analysts perform better over longer horizons.
\end{enumerate}

Our findings highlight the significant potential of LLMs to provide accurate and interpretable predictions for stock ratings. Future work will focus on using longer windows for news summaries, summarizing over extended periods to provide a more comprehensive context. Additionally, we will further explore the capability of LLMs in short-term predictions and develop strategies to enhance their long-term forecasting abilities.

\bibliographystyle{ACM-Reference-Format}
\bibliography{main}

\end{document}